\pdfoutput=1

\documentclass[11pt]{article}

\usepackage[]{emnlp2021}

\usepackage{times}
\usepackage{latexsym}
\usepackage{graphicx}
\usepackage[T1]{fontenc}

\usepackage[utf8]{inputenc}
\usepackage{amsmath}

\usepackage{microtype}

\title{fBERT: A Neural Transformer for Identifying Offensive Content}

\author{Diptanu Sarkar\textsuperscript{1}, Marcos Zampieri\textsuperscript{1}, Tharindu Ranasinghe\textsuperscript{2}, Alexander Ororbia\textsuperscript{1} \\
  \textsuperscript{1}Rochester Institute of Technology, USA \\
  \textsuperscript{2}University of Wolverhampton, UK \\
  \texttt{ds9297@rit.edu} \\
  }

\begin{document}
\maketitle
\begin{abstract}
Transformer-based models such as BERT, XLNET, and XLM-R have achieved state-of-the-art performance across various NLP tasks including the identification of offensive language and hate speech, an important problem in social media. In this paper, we present fBERT, a BERT model retrained on SOLID, the largest English offensive language identification corpus available with over $1.4$ million offensive instances. We evaluate fBERT's performance on identifying offensive content on multiple English datasets and we test several thresholds for selecting instances from SOLID. The fBERT model will be made freely available to the community. 
\end{abstract}

\section{Introduction}
\label{sec:intro}

To cope with the spread of offensive content and hate speech online, researchers have worked to develop automatic methods to detect such posts automatically. Early efforts included methods that used various linguistic features in tandem with linear classifiers \cite{malmasi2017} while, more recently, deep neural networks (DNNs) \cite{ranasinghe2019brums}, transfer learning \cite{wiedemann-etal-2020-uhh, abu-farha-magdy-2020-multitask}, and pre-trained language models \cite{liu2019nuli,ranasinghe-etal-2020-multilingual,ranasinghe2021mudes} have led to even further advances. As evidenced in recent competitions, the performance of these models varies with the sub-task that they are designed to address as well as the datasets used to train them. For example, classical statistical learning models such as the support vector machine (SVM) have outperformed neural transformers in hate speech detection at HatEval 2019 \cite{basile-etal-2019-semeval} and in aggression detection at TRAC 2018 \cite{kumar-etal-2018-benchmarking}. However, for both of these tasks in OffensEval 2019 and 2020 \cite{offenseval, zampieri-etal-2020-semeval}, which focused on the identification of more general offensive language identification, pre-trained transformer-based models such as BERT \cite{devlin2019bert} outperformed other neural architectures and statistical learning methods. 

The introduction of representations learned through the bidirectional encoding inherent to neural transformers (BERT) \cite{devlin2019bert} has been driven much progress in areas in NLP such as language understanding, named entity recognition, and text classification. The base model is pre-trained on a large English corpus, e.g., Wikipedia, BookCorpus \cite{Zhu_2015_ICCV}, using unsupervised masked language modeling and next sentence prediction objectives to adjust the model weights. Various other transformer-based models have also been introduced including RoBERTa \cite{Liu2019RoBERTaAR}, XLNet \cite{NEURIPS2019_dc6a7e65}, XLM-R \cite{xlmroberta}. All of these models, however, are trained on general-purpose corpora for better language understanding, generally lacking domain-specific knowledge. To cope with this limitation, more recently, domain-specific models have been trained and/or fine tuned to different domains such as finance (FinBERT) \cite{FinBERT2019}, law (LEGAL-BERT) \cite{legalbert2020}, scientific texts (SciBERT) \cite{beltagy2019scibert}, and microblogging (BerTweet) \cite{bertweet}. 

\citet{caselli2021hatebert} recently released HateBERT, a BERT transformer model for abusive language detection trained on the Reddit Abusive Language English dataset (RAL-E). HateBERT achieves competitive performance on a few benchmark datasets but relies heavily on manually annotated labels. Moreover, HateBERT was trained on a task-specific dataset (aggression) instead of a more general dataset that encompasses multiple types of offensive language (e.g. hate speech, cyberbullying, profanity) like the popular OLID \cite{zampieri-etal-2019-predicting} used in OfffensEval 2019 at SemEval. 

To address this gap, in this study, we present fBERT, a pre-trained BERT model trained on SOLID \cite{rosenthal2020largescale}, a recently released large dataset crated using OLID's general annotation model but using semi-supervised learning instead of manually annotated labels. SOLID contains over $1.4$ million English tweets with offensive scores greater than $0.5$. We show that the proposed fBERT outperforms both a plain BERT implementation and HateBERT on various offensive and hate speech detection tasks. 

The contributions of this paper are as follows:
\begin{enumerate}
    \item An empirical evaluation of transformer-based, semi-supervised learning techniques applied to offensive language identification with the clear potential application to many other text classification tasks. 
    \item A comprehensive evaluation of several BERT-based strategies and data selection thresholds for offensive language identification across multiple datasets. 
    \item The release of fBERT, a high-performing, state-of-the-art pre-trained model for offensive language identification.
\end{enumerate}

\section{Related Work}
\label{sec:RW}

The use of large pre-trained transformer models has become widespread in NLP. This includes several recently developed offensive language identification systems based on transformer architectures such as BERT \cite{devlin2019bert}. These systems have achieved top performance in popular competitions such as HASOC 2019 \cite{hasoc2019}, HatEval 2019 \cite{basile-etal-2019-semeval}, OffensEval 2019 and 2020 \cite{offenseval,zampieri-etal-2020-semeval}, and TRAC 2020 \cite{trac2020}. The great performance obtained by these systems provides further evidence that pre-trained transformer models are a good fit for the kind of semantic understanding required when identifying offensive content online.

Most of the top systems submitted to the aforementioned competitions \cite{ranasinghe2019brums, wiedemann-etal-2020-uhh, liu2019nuli}, however, use models pre-trained on standard contemporary texts. User generated content and offensive language online, however, contain its own set of distinctive features that models trained on standard texts may fail to represent. Therefore, fine-tuning pre-trained models to this challenging domain is a promising but under explored research direction. To the best of our knowledge, a recent first attempt to fine-tune a BERT model to deal with offensive language online, HateBERT \cite{caselli2021hatebert}, shows promising results for English on multiple datasets. In this paper, we address some of the limitations of HateBERT, discussed in the introduction of this paper, and present fBERT, a new BERT-based offensive language model made freely available to the research community.

\section{Data}
\label{sec:data}

The limited size of existing datasets has been a bottleneck for offensive language identification. OLID \cite{zampieri-etal-2019-predicting}, the dataset used in OffensEval 2019 and arguably the most popular dataset for this task, contains only $14,100$ tweets. OLID is annotated using a hierarchical annotation taxonomy and, as a result, only a sub-set of the corpus is annotated in the lower levels of the taxonomy, i.e., only a few hundred instances. 

More recently, following the OLID taxonomy, \citet{rosenthal2020largescale} released a large-scale offensive language identification dataset (SOLID) with over $9$ million English tweets. The data is collected using the Twitter streaming API. The annotations include labels learned using semi-supervised methods. One important difference between SOLID and OLID is that SOLID is collected using random seeds, which has been shown to decrease topic bias compared to the target keywords used in OLID. All the usernames and URLs are replaced with placeholders and tweets with less than two words or $18$ characters were discarded. 

For retraining fBERT, we have selected over $1.4$ million offensive instances from SOLID. We considered multiple offensive score thresholds from the SOLID dataset including all instances with scores between $0.5$ and $1.0$ arranged in five bins with $0.1$ increments. The number of instances in each range is available in Table \ref{tab:data}. We did not consider the range $0.9 - 1.0$ given that only a very small number of instances ($2,771$) were in this bin . 

\begin{table}[!ht]
\centering
\begin{tabular}{cc}
\hline
\textbf{Threshold} & \textbf{Instances} \\ \hline
0.5 - 1.0 & 1,446,580 \\ 
0.6 - 1.0 & 1,040,525 \\ 
0.7 - 1.0 & 700,719 \\ 
0.8 - 1.0 & 348,038 \\ 
0.9 - 1.0 & 2,771 \\ \hline
\end{tabular}
\caption{Offensive instances from the SOLID dataset, organized according to threshold.}
\label{tab:data}
\end{table}

\section{Model Architecture}
\label{sec:architecture}

\subsection{Input Representation}
\label{sec:input}
We take the sentence input and tokenize it using WordPiece embeddings \cite{wordpiece} with a $30,000$ token vocabulary as described in \cite{devlin2019bert}. The tokenized input is represented as:
\begin{align}
    \mathbf{X} = (\mathbf{x}_{[CLS]}, \mathbf{x}_{1}, \mathbf{x}_{2},..., \mathbf{x}_{n}, \mathbf{x}_{[SEP]})
\end{align}
where $\mathbf{x}_t$ is the $V$-dimensional one-hot encoding of the $t$-th token in a sequence of $n$ symbols (vocabulary of size $V$).
\noindent The tokenized input is then processed via $Bert(\mathbf{X})$ to generate contextualized embeddings as follows: 
\begin{align}
    \mathbf{H} &= Bert(\mathbf{X})\\
    \mathbf{H} &= (\mathbf{h}_{[CLS]}, \mathbf{h}_{1}, \mathbf{h}_{2}, ..., \mathbf{h}_{n}, \mathbf{h}_{[SEP]})\label{eqn:bert}
\end{align}
where $\mathbf{h}_t$ is the $d$-dimensional embedding for the $t$-th token $\mathbf{x}_t$ (resulting in $n$ embeddings).

\subsection{Retraining Procedure}
\label{sec:retraining}
The goal of the study is to adapt the BERT model for social media aggression detection tasks. We utilized a BERT base (uncased) model that consists of $12$ bidirectional transformers encoders with $768$ hidden layers and $12$ self-attention heads. To use the general understanding of the English language and context, we initialize the transformer with pre-trained weights\footnote{BERT Pre-trained weights: \url{https://github.com/google-research/bert}}. We used over $1.4$ million offensive texts from the SOLID dataset to retrain the model. No cleaning was applied to preserve the incoherent composition of social media posts, such as the excessive use of mentions, emojis, or hashtags. We retrained the model using the masked language modeling objective to adapt the bidirectional representations of social media offensive language.

\begin{figure}[!ht]
\centering
\includegraphics[scale=0.7]{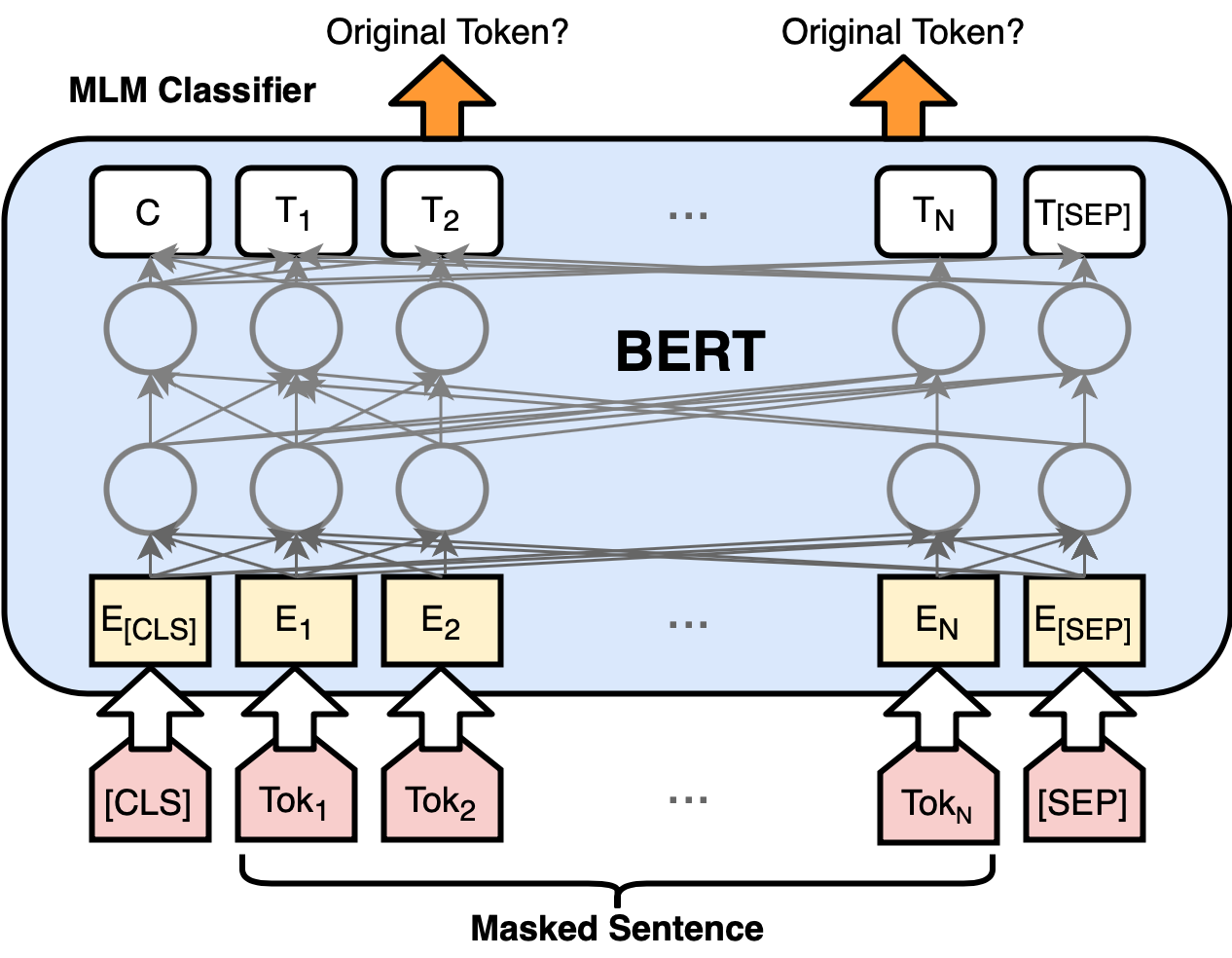}
\caption{A schematic representation of the fBERT masked language model (the re-trained/tuned BERT).}
\label{fig:fbert}
\end{figure}

\paragraph{Masked Language Modeling (MLM)}
\label{sec:mlm}
In MLM, we randomly mask a percentage of tokens and predict the masked inputs. As prescribed in the original BERT implementation, we randomly select $15\%$ of the total tokens for replacement, $80\%$ of the selected tokens are replaced with $[MASK]$, $10\%$ are substituted with a random token chosen from the vocabulary, and $10\%$ remain unchanged. The hidden vectors with masked tokens are fed into a softmax activation function to generate a probability distribution over each (masked) token $\mathbf{x}_t$:
\begin{align}
    p(\mathbf{x}_t | \mathbf{H}) = softmax(\mathbf{W} \cdot \mathbf{h}_t + \mathbf{b})
\end{align}
where $\cdot$ is matrix multiplication, $\mathbf{W} \in \mathcal{R}^{d \times V}$, and $\mathbf{b} \in \mathcal{R}^{V \times 1}$.
\noindent The model is trained to predict the original token by minimizing the Catergorical cross-entropy objective as follows:
\begin{align}
    \mathcal{L} = -\sum^n_{t=1} m_t \sum_v \Big( \mathbf{x}_t \otimes \log( p(\mathbf{x}_t|\mathbf{H}) ) \Big)[v]
\end{align}
where $m_t$ is the binary scalar applied at time step $t$ ($1$ if the word is masked, $0$ otherwise). $[v]$ retrieves/indexes the $v$th item in the vector $\mathbf{x} \otimes \log(p(\mathbf{x}|\mathbf{H})$ and $\otimes$ indicates element-wise multiplication.
 A schematic representation of the BERT masked language model is presented in Figure \ref{fig:fbert}.

\paragraph{Retraining Setup}
\label{sec:setup}
We trained the resulting fBERT for $25$ epochs using the MLM objective with $0.15$ probability to randomly mask tokens in the input. The language model is trained with a batch size of $32$ and a $512$ maximum token length using the Adam optimizer with a learning rate of $5e-5$. The training time took $5$ days on a single Nvidia V$100$ GPU. 

\section{Experiments}
\label{sec:fbert_exp}

To determine the effectiveness and portability of the trained fBERT, we conducted a series of experiments using benchmark datasets and compared our model with a general-purpose BERT model. We used the same set of configurations for all the datasets evaluated in order to ensure consistency between all the experiments. This also provides a good starting configuration for researchers who intend to use fBERT on a new dataset.

We used a batch-size of eight, Adam optimiser with learning rate $1\mathrm{e}{-4}$, and a linear learning rate warm-up over 10\% of the training data. During the training process, the parameters of the transformer model, as well as the parameters of the subsequent layers, were updated. The models were trained using only training data. Furthermore, they were evaluated while training using an evaluation set that had one fifth of the rows in training data. We performed early stopping if the evaluation loss did not improve over ten evaluation steps. All the models were trained for three epochs. The rest of the parameters are shown in Table \ref{tab:parameter}. These experiments were also conducted in a Nvidia V$100$ GPU.

\begin{table}[!ht]
\centering
\setlength{\tabcolsep}{4.5pt}
\scalebox{0.95}{
\begin{tabular}{ll}
\hline
\bf Parameter & \bf Value  \\ \hline
learning rate & 1e-4 \\
adam epsilon & 1e-8       \\
warmup ratio & 0.1    \\
warmup steps  & 0       \\
max grad norm & 1.0        \\
max seq. length & 140        \\
gradient accumulation steps & 1 \\
 \hline
\end{tabular}
}
\caption{Parameter Specifications.}
\label{tab:parameter}
\end{table}

\paragraph{HatEval 2019} \label{para:hateval} In the SemEval 2019, HatEval \cite{basile-etal-2019-semeval} introduced the challenge of detecting multilingual hate speech against women and immigrants. The dataset for the task is collected from Twitter in both English and Spanish. In this work, we used only the English dataset comprised of $9,000$ training instances with $4,177$ hateful tweets. The development (dev) and test sets contain $1,000$ ($123$ instances are hateful) and $3,000$ examples ($1,380$ instances are hateful) respectively. In terms of pre-processing, we removed extra whitespaces, usernames and URLs were replaced with placeholders, the Emoji\footnote{Emoji Package: \url{https://pypi.org/project/emoji/}} package was used to convert the emojis to text, and the Word Segmentation\footnote{Word Segmentation Package: \url{https://pypi.org/project/wordsegmentation/}} package was used to segment the words into hashtags. We applied the same pre-processing steps for all models to compare the test set macro F$_1$ score.

\paragraph{OLID} We use OLID, the official dataset for OffensEval 2019 \cite{offenseval}, one of the the most popular offensive language identification shared tasks. The dataset has $13,240$ training and $860$ test instances. There are $4,400$ and $240$ offensive posts in the training and test dataset, respectively. For the experiment, we chose sub-task A, a binary classification task between offensive and non-offensive posts. We used $10\%$ of the training data as development data and performed pre-processing and cleaning steps as described by  \citet{liu2019nuli}. We trained fBERT for the offensive language detection task and compared its performance with other language models using the macro F$_1$ score.


\paragraph{Hate Speech and Offensive Language Detection (HS \& O)} In fine-grain aggression detection, classifying offensive language and hate speech is challenging. Hate speech contains explicit instances targeted towards a specific group of people intended to degrade or insult. \citet{davidson2017} compiled a $24,783$ English tweets dataset annotated with one of three labels -- ``hate speech'', ``only offensive'', and ``neither''. The dataset contains $1,430$ hate speech, $19,190$ only offensive, and $4,163$ instances that are neither. We further split the dataset into training, dev, and test sets in a 3:1:1 ratio. We applied the same preprocessing steps we applied to the HatEval 2019 dataset.

\section{Results}

We first present the results for the SOLID data selection thresholds in Table \ref{table:thresholds} in terms of F$_1$ Macro. For the three datasets tested, the 0.5 - 1.0 threshold, which includes the largest number of instances, yielded the best performance. 

\begin{table}[!ht]
\centering
\begin{tabular}{c|ccc}
\hline
& \multicolumn{3}{|c}{\textbf{Datasets}} \\ 
\bf Scores     & \bf HatEval & \bf OLID & \bf HS \& O \\ \hline
0.5 - 1.0 & \bf 0.596  & \bf 0.813     & \bf 0.878       \\
0.6 - 1.0 & 0.562  & 0.808     & 0.871    \\
0.7 - 1.0 & 0.550  & 0.802     & 0.867      \\
0.8 - 1.0 & 0.554  & 0.801     & 0.865      \\ \hline
\end{tabular}
\caption{Macro F$_1$ scores for different SOLID threshold score values.}
\label{table:thresholds}
\end{table}

\noindent We then compare the performance of fBERT with BERT and HateBERT. In the HatEval Sub-task A, we see that fBERT has outperformed BERT by increasing the test macro F$_1$ score by over $23\%$. This empirically demonstrates the advantage and generalization power of our domain-specific retrained BERT model. The best model \cite{indurthi-etal-2019-fermi} in this task used an SVM model with a radial basis kernel, exploiting sentence embeddings from Google's Universal Sentence Encoder as features. The results are shown in Table \ref{table:fbert_oe}. 

The fBERT model also performs better than the generic BERT and abusive language HateBERT in OffensEval Sub-task A, achieving a test set macro F$_1$ score of $0.8132$. We observe that the fBERT is also highly effective in fine-grain offensive and hate speech detection, obtaining a $10\%$ increase in the F$_1$ score. 

\begin{table}[!ht]
\centering
\begin{tabular}{llc}
\hline
\bf Dataset & \textbf{Model} & \textbf{Macro F1} \\ \hline
& \bf fBERT & \bf 0.596 \\ \
HatEval & HateBERT & 0.525 \\ 
& BERT & 0.483 \\ \hline

& \bf fBERT & \bf 0.813 \\ 
 OLID & HateBERT & 0.801 \\ 
& BERT & 0.794 \\ \hline

& \bf fBERT & \bf 0.878 \\ 
HS \& O & HateBERT & 0.846 \\
& BERT & 0.806 \\ \hline

\end{tabular}
\caption{The test set macro F$_1$ scores for all datasets and models. Results are ordered by performance. Best results are shown in bold font.}
\label{table:fbert_oe}
\end{table}

\noindent Finally, as observed in the experimental results presented above, we observe that fBERT has outperformed the abusive language HateBERT model in all of the experiments. The proposed fBERT has also performed efficiently in all the aggression detection tasks. This validates the effectiveness of the proposed domain-specific transformer model for offensive and hateful language classification tasks. The proposed fBERT model is effective across different datasets and objectives, providing a powerful model to be used for hateful/offensive content identification.

\section{Conclusion}

Over the years, neural transformer models have outperformed previous state-of-the-art deep learning models across various NLP tasks including offensive and hate speech detection tasks. Nevertheless, these transformers are usually trained on general corpora which lack tweet and offensive language-specific cues. Previous studies have shown that domain-specific fine-tuning or retraining of models before attempting to solve downstream tasks can lead to excellent results in multiple domains. As discussed in this paper, fine-tuning/retraining a complex models to identify offensive language has not been substantially explored before and we address this gap by proposing fBERT, a \textit{bert-base-uncased} model that has been learned using over $1.4$ million offensive instances from the SOLID dataset. The shifted fBERT model better incorporates domain-specific offensive language and social media features. The fBERT model achieves better results in both OffensEval and HatEval tasks and in the HS \& O dataset over BERT and HateBERT. 

In future work, we would like to investigate the performance of fBERT both at the post- and token-level identification stages. Furthermore, we will expand fBERT to multiple languages. Since our approach is based on a semi-supervised dataset, it is easily expandable to other languages as well. We plan to extend this process to other transformer models such as XLNET \cite{NEURIPS2019_dc6a7e65}, RoBERTa \cite{Liu2019RoBERTaAR} and ALBERT \cite{Lan2020ALBERT}. Finally, fBERT is publicly available on Hugging Face model hub \cite{wolf-etal-2020-transformers}.\footnote{fBERT at HuggingFace: \url{https://huggingface.co/diptanu/fBERT}}

\section*{Ethics Statement}

fBERT is essentially a BERT model for offensive language identification which is trained on multiple publicly available datasets. We used multiple datasets referenced in this paper which were previously collected and annotated. No new data collection has been carried out as part of this work. We have not collected or processed writers'/users' information nor have we carried out any form of user profiling protecting users' privacy and identity. 

We understand that every dataset is subject to intrinsic bias and that computational models will inevitably learn biased information from any dataset. We believe that fBERT will help coping with biases in datasets and models as it features a freely available BERT model that other researchers can use to train new offensive language identification models on other datasets.

\section*{Acknowledgments}
We would like to thank the HatEval and OffensEval shared task organizers for making the datasets used in this paper available. We further thank the anonymous EMNLP reviewers for their insightful feedback. 

This research was partially supported by a seed fund award sponsored by RIT's Global Cybersecurity Institute (GCI). 

\bibliography{emnlp}
\bibliographystyle{acl_natbib}

\end{document}